\documentclass[letterpaper, 10 pt, conference]{ieeeconf}  

\IEEEoverridecommandlockouts                              

\usepackage{graphics} 
\usepackage{epsfig} 
\usepackage{mathptmx} 
\usepackage{times} 
\usepackage{amsmath} 
\usepackage{amssymb}  
\usepackage{tabularx}
\usepackage{color,soul}
\usepackage{multirow}
\usepackage{hyperref}
\usepackage[font=small]{caption}
\usepackage{float}

\title{\LARGE \bf
A Learning Approach to Robot-Agnostic Force-Guided \\ High Precision Assembly 
}

\author{Jieliang Luo* and Hui Li*
\thanks{*The authors contributed equally and are with Autodesk Research, San Francisco, United States. Emails: {\tt\small rodger.luo@autodesk.com}, {\tt\small hui.xylo.li@autodesk.com}}%
}

\begin{document}

\maketitle
\thispagestyle{empty}
\pagestyle{empty}

\begin{abstract}

In this work we propose a learning approach to high-precision robotic assembly problems. We focus on the contact-rich phase, where the assembly pieces are in close contact with each other. Unlike many learning-based approaches that heavily rely on vision or spatial tracking, our approach takes force/torque in task space as the only observation. Our training environment is \emph{robotless}, as the end-effector is not attached to any specific robot. Trained policies can then be applied to different robotic arms without re-training. This approach can greatly reduce complexity to perform contact-rich robotic assembly in the real world, especially in unstructured settings such as in architectural construction. To achieve it, we have developed a new distributed RL agent, named Recurrent Distributed DDPG (RD2), which extends Ape-X DDPG\cite{horgan2018distributed}  with recurrency and makes two structural improvements on prioritized experience replay\cite{schaul2015prioritized}. Our results show that RD2 is able to solve two fundamental high-precision assembly tasks, lap-joint and peg-in-hole, and outperforms two state-of-the-art algorithms, Ape-X DDPG and PPO with LSTM. We have successfully evaluated our robot-agnostic policies on three robotic arms, Kuka KR60, Franka Panda, and UR10, in simulation. The video presenting our experiments is available at \url{https://sites.google.com/view/rd2-rl} 

\end{abstract}

\section{Introduction}

\begin{figure*}[h]
    \centering
    \includegraphics[width=1\textwidth]{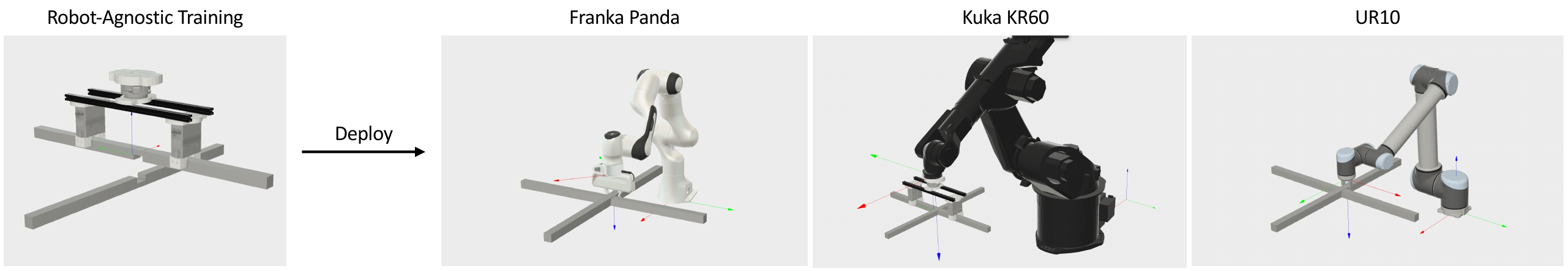}
    \caption{We develop a learning approach to solving robotic assembly tasks in the contact-rich phase with F/T measurements as the only observation. The training environment in simulation is robot-agnostic so that the polices can be deployed on various robotic arms.}
    \label{fig:teaser}
\end{figure*}

Robotic systems for automated assembly have been widely used in manufacturing, where the environment can be carefully and precisely controlled, but they are still in infancy in architectural construction. A main reason is that current robotic systems are not adaptive to the diversity of the real world, especially in unstructured settings. RL-based robotic systems~\cite{zhu2020ingredients, andrychowicz2020learning, ren2018learning} are a promising direction given their adaptability to uncertainties. Current successful RL examples heavily rely on vision or spatial tracking to perform complex control tasks~\cite{schoettler2019deep, akkaya2019solving,luo2020dynamic, beltran2020variable}. However, it is unrealistic to expect motion capture or other tracking systems at a construction site, as they are hard to install, calibrate, and scale. A vision system is more portable, but in the contact-rich phase of assembly, it often fails to help due to occlusion or poor lighting conditions. Another limitation of the current RL-based robotic systems is that the policies are robot-specific and cannot readily generalize to other robotic platforms. This limits the efficiency and scalability of RL-based robotic systems, especially in construction, where often a collection of different robotic platforms are used.

In this paper, we present a learning approach to solving high-precision robotic assembly tasks of varying complexity. We are interested in the contact-rich phase of assembly, because when the assembly pieces are in close contact with each other, force/torque (F/T) measurements become the most revealing observation. The research question we want to answer is: in this contact-rich phase, is F/T alone sufficient for learning a robot control policy? To do this, we develop a recurrent distributed deep RL agent called Recurrent Distributed DDPG (RD2) that extends Ape-X DDPG and solves the force-guided robotic assembly problems in the continuous action space. Specifically, we add recurrency in the neural networks to learn a memory-based representation in order to compensate for partial observability of the process, i.e. the lack of pose observations. Additionally, we create a dynamic scheme to pre-process episode transitions before sending them to the reply buffer by allowing overlap of the last two sequences in each episode to be variable, which maintains important information of the final transitions and avoids crossing episode boundaries. To overcome the training instability of the DDPG family~\cite{barth2018distributed, horgan2018distributed, lillicrap2016ddpg, heess2015memory}, we calculate priorities for the sequences in the replay buffer as well as priorities of the transitions in each sequence. The former priorities are used for sequence sampling and the latter are for bias annealing.  We use \textit{robotless} environments for training. During the contact-rich phase of assembly, our tasks can be easily confined within the robot workspace that is collision free and singularity free, and robot motion can be considered quasi-static. This ensures the policies can successfully transfer to various robotic arms without re-training.  

We evaluate RD2 on two fundamental assembly tasks, lap-joint and peg-in-hole, with tight tolerance and varying complexity, and compare the performance of RD2 to Ape-X DDPG and PPO~\cite{schulman2017proximal} with LSTM (LSTM-PPO). The results show that RD2 outperforms the other two algorithms across all the tasks and presents a stable performance as the difficulty of the tasks increases, whereas Ape-X DDPG and LSTM-PPO fail on most of the tasks. We also show that the trained robotless policies adapt well to different robotic arms with different initial states and different physical noise injected in F/T measurements and friction parameters in simulation. 

The main contribution of this work is a robot-agnostic learning approach to solving contact-rich high-precision robotic assembly tasks with F/T as the only observation. Trained polices can be deployed on different robotic arms without re-training. We believe this work is an important step towards deploying robots on unstructured construction sites, where different robots can share trained policies and adapt to misalignment with minimal sensing setup.



The remainder of this paper is structured as follows. Problem statement and related work are stated in Section~\ref{sec:background}, followed by a detailed explanation of our proposed approach in Section~\ref{sec:method}. Experimental setup, results, and evaluation are presented in Section~\ref{sec:exp}. Section~\ref{sec:conclusion} concludes the paper and discusses future work.

\section{Problem Statement and Related Work}
\label{sec:background}

\subsection{Problem Statement}

We model the problem we solve in this paper as a Partially Observable Markov Decision Process (POMDP), which is described by a set of states $S$, a set of actions $A$, a set of conditional probabilities $p(s_{t+1} | s_t, a_t)$ for the state transition $s_t \rightarrow s_{t+1}$, a reward function $R : S \times A \rightarrow \mathbb{R}$, a set of observations $\Omega$, a set of conditional observation probabilities $p(o_t | s_t)$, and a discount factor $\gamma \in [0, 1]$.    

In principle, the agent makes decisions based on the history of observations and actions $h_t = (o_1, a_1, o_2, a_2 ..., o_t, a_t)$ and the goal of the agent is to learn an optimal policy $\pi_\theta$ in order to maximize the expected discounted rewards:

\begin{equation}
\label{eq1}
\max_{\pi_\theta}\mathbb{E}_{\tau\sim\pi_\theta}\left[ \sum_{t=1}^{T} \gamma^{t-1} r(s_t, a_t) \right],
\end{equation}

where trajectory $\tau = (s_1, o_1, a_1, s_2, o_2, a_2, ..., s_T, o_T, a_T)$, $\theta$ is the parameterization of policy $\pi$, and $\pi_\theta(\tau)=p(s_1)p(o_1 | s_1)\pi_\theta(a_1|h_1)\prod_2^T p(s_t|s_{t-1},a_{t-1}) p(o_t | s_t) \pi_\theta(a_t|h_t)$.

For many POMDP problems, it is not practical to condition on the entire history of the observations~\cite{heess2015memory}. In this paper, we tackle this challenge by investigating recurrent neural networks with distributed model-free RL and we focus on the continuous action domain.

\subsection{Distributed Reinforcement Learning}

Distributed reinforcement learning can greatly improve sample efficiency of model-free RL by decoupling learning and data collection. 

Ape-X~\cite{horgan2018distributed} disconnects exploration from learning by having multiple actors interacting with their own environment and sending the collected transitions into one of the distributed replay buffers. A learner asynchronously samples a batch of transitions from a randomly picked buffer. Ape-X has both DQN~\cite{mnih2015dqn} and DDPG~\cite{lillicrap2016ddpg} variants to support discrete and continuous action spaces, respectively. Built upon Ape-X, D4PG~\cite{barth2018distributed} introduced a distributional critic update and incorporated N-step returns and prioritization of experience replay to achieve a more stable learning signal. 

Unlike the actors and the learner in Ape-X that randomly feed and samples transitions from replay buffers, IMPALA~\cite{espeholt2018impala} asks each actor to send the collected transitions via a first-in-first-out queue to the learner and to update its policy weights from the learner before the next episode. In addition, it introduces V-trace, a general off-policy learning algorithm that corrects a \textit{policy-lag} between the learner and the actors as the actors' policies are usually several updates behind the learner's. Both Ape-X and IMPALA have demonstrated strong performance in the Atari-57 and DMLab-30 benchmarks, but they have not been examined on partially observable robotic assembly tasks.   

\subsection{RL for Robotic Assembly}

RL has been studied actively in the area of robotic assembly as it can reduce human involvement and increase robustness to uncertainty. Dynamic Experience Replay~\cite{luo2020dynamic} uses experience replay samples not only from human demonstrations but also successful transitions generated by RL agents during training and therefore improves training efficiency. ~\cite{fan2018guided} proposed a framework to combine DDPG~\cite{lillicrap2016ddpg} and GPS~\cite{levine2013guided} to take advantage of both model-free and model-based RL~\cite{sutton2018reinforcement} to solve high-precision Lego insertion tasks. \cite{lee2019making} uses self-supervision to learn a multimodal representation of visual and haptic feedback to improve sample efficiency. Both ~\cite{ren2018learning} and ~\cite{beltran2020variable} focus on training a variable compliance controller to solve peg-in-hole tasks. ~\cite{ren2018learning} introduces incremental displacement and force in the observation space and visual Cartesian stiffness in the action space to improve the efficiency of the trained policies. ~\cite{beltran2020variable} applies domain transfer-learning techniques to improve training efficiency and domain randomization to increase the adaptability of the learned policies. \cite{li2021} solves the problem of timber joint assembly in architectural construction. These methods all require pose information, directly or indirectly, as observations. In our work, we focus on the contact-rich phase and only use measurements from the wrist-mounted F/T sensor as observations. This allows the RL system to solve assembly tasks where vision or pose tracking systems are unavailable, e.g. inside a small confined space.

\begin{figure*}[h]
    \centering
    \includegraphics[width=0.71\textwidth]{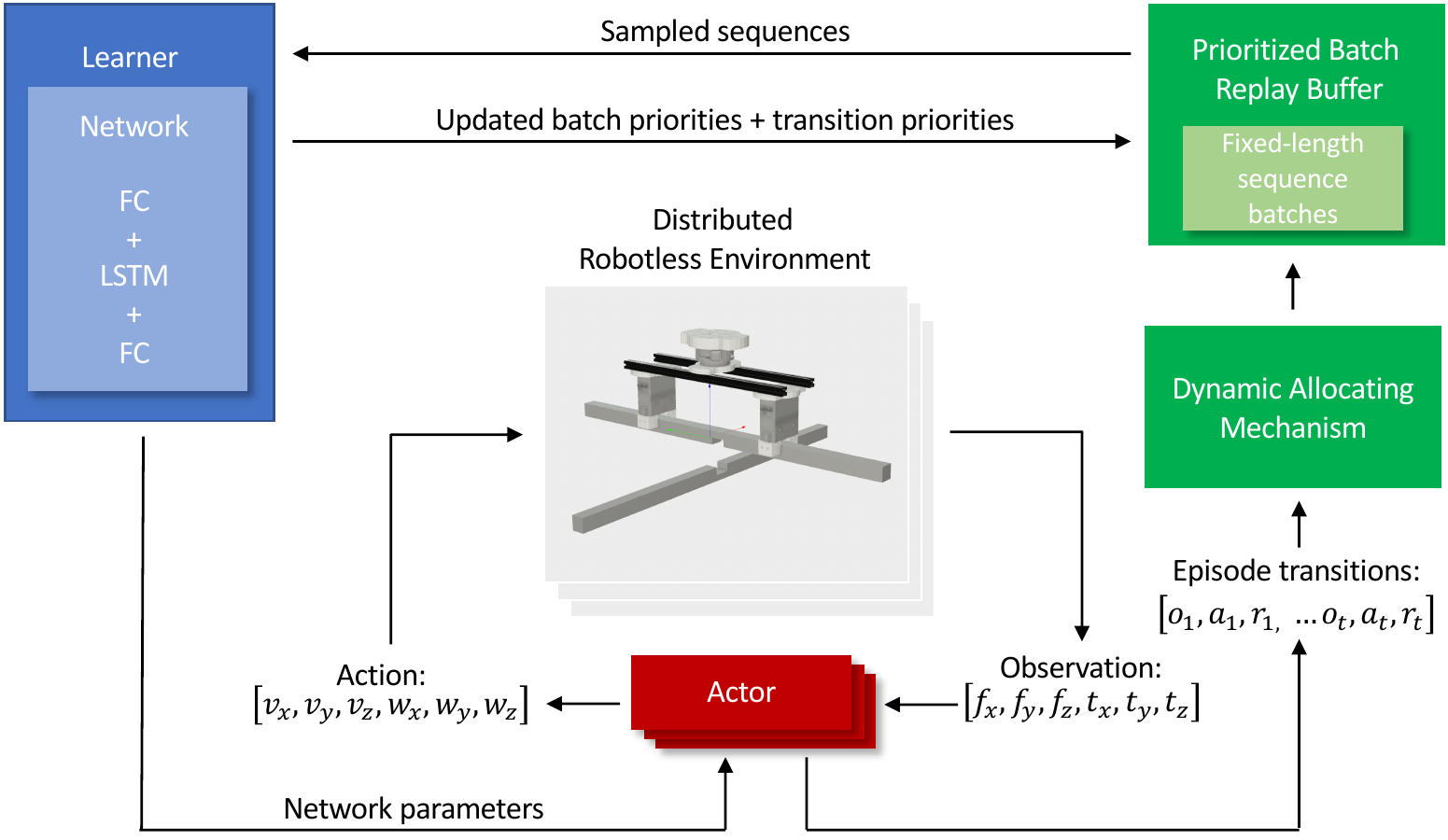}
    \caption{The architecture of RD2: it has multiple actors, each with its own instance of environment, to collect episodes of $(o, a, r)$. Since the length of each episode is uncertain, we design a dynamic allocating mechanism to break one episode of transitions $(o_1, a_1, r_1, ..., o_t, a_t, r_t)$ into a group of fixed-length sequences of transitions, saved in a prioritized batch replay buffer. A learner samples the sequences based on their priorities, updates its networks, and sends the priorities of the sequences and the priorities of each individual transition back to the buffer. The actors periodically update their networks' parameters from the learner.}
    \label{fig:rd2-architecture}
\end{figure*}

\subsection{RL under Partial Observability}

Partial observability is a well known challenge in robotics, resulting from occlusions, unpredictable dynamics, or noisy sensors. \cite{hausknecht2015deep} proposed RDQN, which replaces the first post-convolutional fully-connected layer with an LSTM layer in DQN. This modification allows the agent to only see a single frame at each timestep, but is capable of replicating DQN's performance on standard Atari games. Built upon Ape-X DQN, \cite{kapturowski2018recurrent} extended RDQN to R2D2, where LSTM-based agents learn from distributed prioritized experience replay. The result shows that the R2D2 agent can unprecedentedly exceed human-level performance in 52 of the 57 Atari games. Because the action space of the two algorithms is discrete, they cannot address continuous control problems in robotics.   

On the robotic side, \cite{openai2018learning} used LSTM as an additional hidden layer in PPO to train a five-fingered humanoid hand to manipulate a block. The memory-augmented method was a key factor in successfully transferring the policy trained in randomized simulations to a real robotic hand, suggesting that the use of memory could help the policy to adapt to a new environment. \cite{inoue2017deep} used a Q-learning based method with two LSTM layers for Q-function approximation to solve low-tolerance peg-in-hole tasks. Although memory-augmented policies have proven to improve training results in continuous control problems, neither method investigated in partially observable tasks with minimal set of observations.

\section{Method}

\label{sec:method}
In this section, we introduce the setup of the RL training environment, explain the details of the RD2 agent, and describe our method to transfer robot-agnostic polices to robotic arms.

\subsection{Setup}
For both training and deployment, we use an internal simulator with the Bullet~\cite{coumans2016pybullet} physics engine. 

\textbf{Observation:} The observation space is 6-dimensional, being the F/T measurements ($f_x,f_y,f_z,\tau_x,\tau_y,\tau_z$) from the sensor, which is mounted at the robot end-effector. 

\textbf{Action:} The action space is continuous and 6-dimensional, being the desired Cartesian-space linear velocity ($v_x,v_y,v_z$) and angular velocity ($w_x,w_y,w_z$) at the center of the assembly piece under control.  

\textbf{Reward:} We use a simple linear reward function based on the distance between the goal pose and the current pose of the moving joint member. Additionally we use a large positive reward (+100) if the current pose is within a small threshold of the goal pose: 

\[ 
r= \left \{
  \begin{tabular}{ccc}
  $-| g - x |$, & $| g - x | > \epsilon$ \\
  $-| g - x | + R$, & $| g - x | \leq \epsilon$ 
  \end{tabular}
  \right.
\]

where $x$ is the current pose of the joint member, $g$ is the goal pose, $\epsilon$ is the distance threshold, and $R$ is the large positive reward. We use the negative distance as our reward function to discourage the behavior of loitering around the goal because the negative distance also contains time penalty. 

Note that the reward function is only used during training in simulation, where distance is easy to acquire, and is not used during rollouts. Hence, no vision or spatial tracking system is needed when deploying the policies on real robots. 

\textbf{Termination: } An episode is terminated when the distance between the goal pose and the pose of the joint member is within a pre-defined threshold or when a pre-defined number of timesteps are reached. 

\subsection{The Recurrent Distributed DDPG Agent}
\label{sec:agent}

We propose Recurrent Distributed DDPG (RD2) to solve the partially observable assembly tasks. Built upon Ape-X DDPG, RD2 adds an LSTM layer between the first fully-connected layer and the second fully-connected layer in both the actor and the critic networks as well as their target networks. No convolutional layers are added since we explore in the low-dimensional observation space. The details of the network architecture is provided in Table~\ref{table: nn-structure}. 

\begin{table}[h]
\setlength{\arrayrulewidth}{0.1mm}
\setlength{\tabcolsep}{5pt}
\renewcommand{\arraystretch}{1.2}
\centering
\begin{tabular}{|c|c|c|}
 \hline
 \textbf{Layer} & \textbf{Input Size} & \textbf{Output Size}\\
 \hline
 Fully-connected (Relu) & [\#Observations] & [256] \\
 \hline
 LSTM (Relu) & [256] & [256]\\
\hline
Fully-connected (Q Network) & [256] & [1]\\
 \hline
Fully-connected (Tanh) ($\pi$ Network) & [256] & [\#Actions]\\
\hline
\end{tabular}
\caption{Specification of the Q network and $\pi$ network used in RD2 for all the experiments. It also applies to their target networks.}
\label{table: nn-structure}
\end{table}

In the experience replay buffer, we store fixed-length sequences of transitions. Each sequence contains ($m$, $m$=2$k$, where $k\in\mathbb{Z^+}$) transitions, each of the form $(observation, action, reward)$. Adjacent sequences overlap by $m/2$ timesteps and the batches of sequences never cross the episode boundary. As the length of each episode in the assembly tasks varies, which may result in some sequences containing transitions from two episodes if we naively segment transitions into fixed-length sequences, we introduce a dynamic mechanism to allow the last overlap in each episode to be a variable between $[m/2, m-1]$. Specifically, the last overlap is calculated as:

\[ 
O = \left \{
  \begin{tabular}{ccc}
  $m - T\bmod{(m/2)}$, & $T\bmod{(m/2)} \neq 0$\\
  $m/2$, & $T\bmod{(m/2)} = 0$\\
  \end{tabular}
  \right.
\]

where $O$ is the number of transitions in the last overlap and $T$ is the total timesteps in each episode. This mechanism prevents losing or compromising any transitions at the end of each episode, which usually contain crucial information for training.    

Similar to R2D2, we sample the sequences in the replay buffer based on their priorities, formulated as:

\[ p = \eta \max(\delta) + (1 - \eta) \bar{\delta}\]

where $\delta$ is a list of absolute n-step TD-errors in one sequence. We set $\eta$ to $0.9$ to avoid compressing the range of priorities and limiting the ability to pick out useful experience.

In addition, as discussed in~\cite{schaul2015prioritized}, prioritized replay introduces bias because it changes the distribution of the stochastic updates in an uncontrolled fashion, and therefore changes the solution that the estimates converge to. For each transition in a sequence, we correct the bias using importance-sampling weights: 

\[
w_i = (N \times P(i))^{-\beta}
\]

where $N$ is the size of the replay buffer and we set $\beta$ to 0.4. We normalize the weight of each transition before sending the sequences for backpropagation through time (BPTT)~\cite{werbos1990backpropagation} by $1/max_iw_i$. On the implementation level, we initialize two sum-tree data structures such that one keeps the priorities of the sequences and the other one keeps the priorities of the transitions. We observe that this step is crucial to stabilize the training process for our tasks. The details of the RD2 architecture is shown in Fig.\ref{fig:rd2-architecture}. 

We use a zero start state in LSTM to initialize the network at the beginning of the sampled sequence and train the RD2 agent with Population Based Training (PBT)~\cite{jaderberg2017population} on an AWS p3.16xlarge instance. Every training session includes 8 concurrent trials, each of which contains one single GPU-based learner and 8 actors. We make the size of the batches, the length of the sequences, and n-step as mutable hyper-parameters to PBT. Each trial evaluates in every 5 iterations to determine whether to keep the current training or copy from a better trial. If a copy happens, the mutable hyper-parameters are perturbed by a factor of 1.2 or 0.8 or have 25\% probability to be re-sampled from the original distribution. The details of the hyper-parameters fine-tuned in PBT are provided in Appendix.

\subsection{Robotless-to-Robot Transfer}
As we focus on the contact-rich phase of assembly, our tasks can be confined within the robot workspace that is collision free and singularity free. Moreover, robot motion in our tasks is slow enough to be considered quasi-static, and hence, the robot dynamics and inertial forces can be ignored. 

In order to transfer policies trained in the robotless environment to the deployment environment of a specific robotic arm, we consider both the observation and the action spaces. For the observation space, we apply a coordinate transformation to the F/T measurements, using the force-torque twist matrix. If we denote F/T from its coordinates in the robotless environment (frame $\mathcal{F}_b$) as ${}^bh=({}^bf, {}^b\tau)$, and denote F/T from its coordinates in the end-effector of a robotic arm (frame $\mathcal{F}_a$) as ${}^ah=({}^af, {}^a\tau)$, then the transformation is shown in Eq.~\ref{eq:grasp}.
\begin{equation}
{}^ah = \begin{bmatrix}\label{eq:grasp}
{}^aR_b & 0_3 \\ [{}^at_b] {}^aR_b & {}^aR_b 
\end{bmatrix} {}^bh
\end{equation}
Where ${}^aR_b$ and ${}^at_b$ are the rotation matrix and the translation vector respectively from frame $\mathcal{F}_a$ to frame $\mathcal{F}_b$.

The action space is defined as the Cartesian-space velocity at the center of the assembly piece under control, which is identical across different robotic arm setups. Hence, there is no transformation needed for actions.

\section{Experiments}
\label{sec:exp}

In this section, we answer the following questions: (1) How does RD2 compare to the state-of-the-art RL algorithms on robotic assembly tasks with the force/torque as the only observation? (2) How does the difficulty of the tasks affect the performance of RD2 and its comparative algorithms? (3) How do the robot-agnostic policies perform on various robotic arms and with different initial poses as well as with noise injected in F/T and friction? 

\begin{figure}[h]
    \centering
    \includegraphics[width=0.5\textwidth]{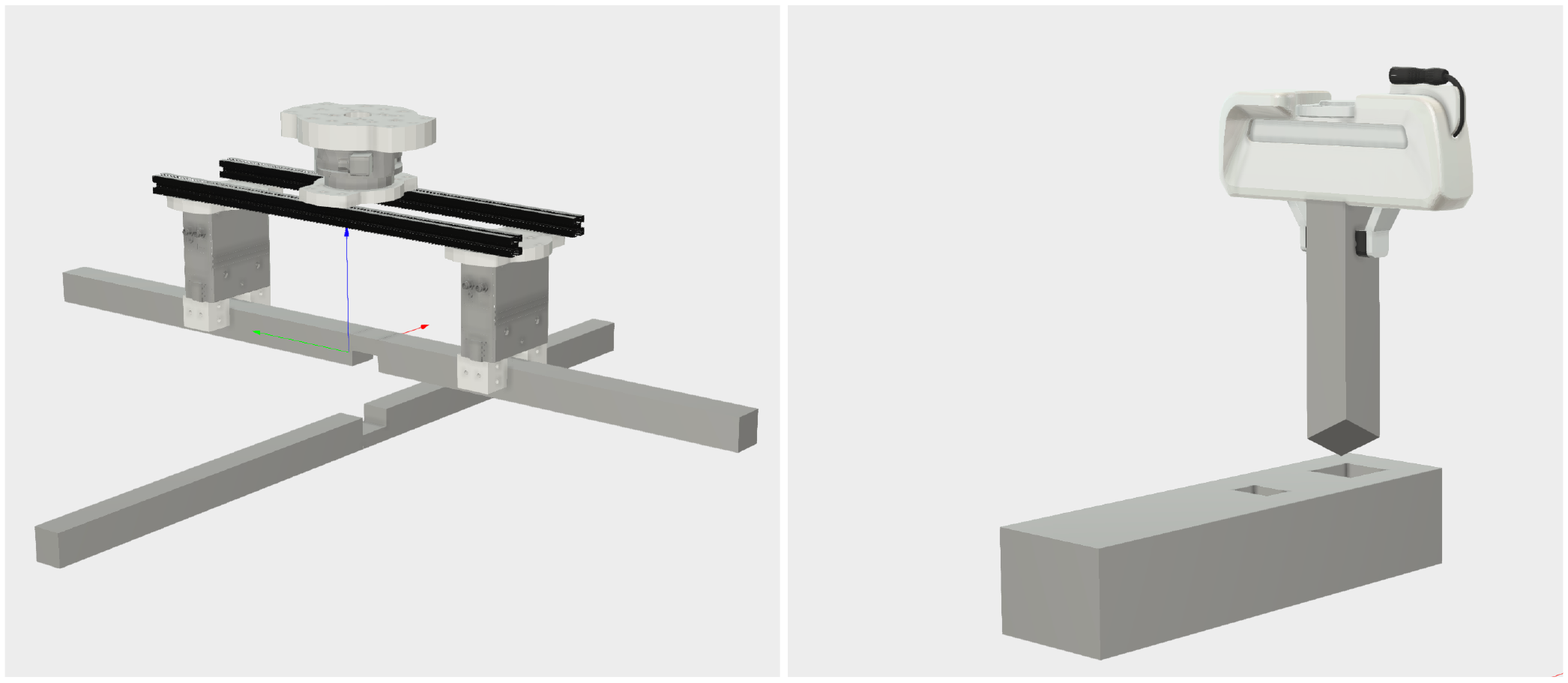}
    \caption{Visualizations of the two robotless assembly environments: the lap-joint environment on the left has 2mm assembly tolerance and the peg-in-hole environment on the right has 0mm tolerance.}
    \label{fig:assembly-env}
\end{figure}

\begin{figure*}[h]
    \centering
    \setlength{\tabcolsep}{0pt}
    \begin{tabular}{ c c c }
         \multirow{2}{*}[3em]{\includegraphics[width=0.31\textwidth]{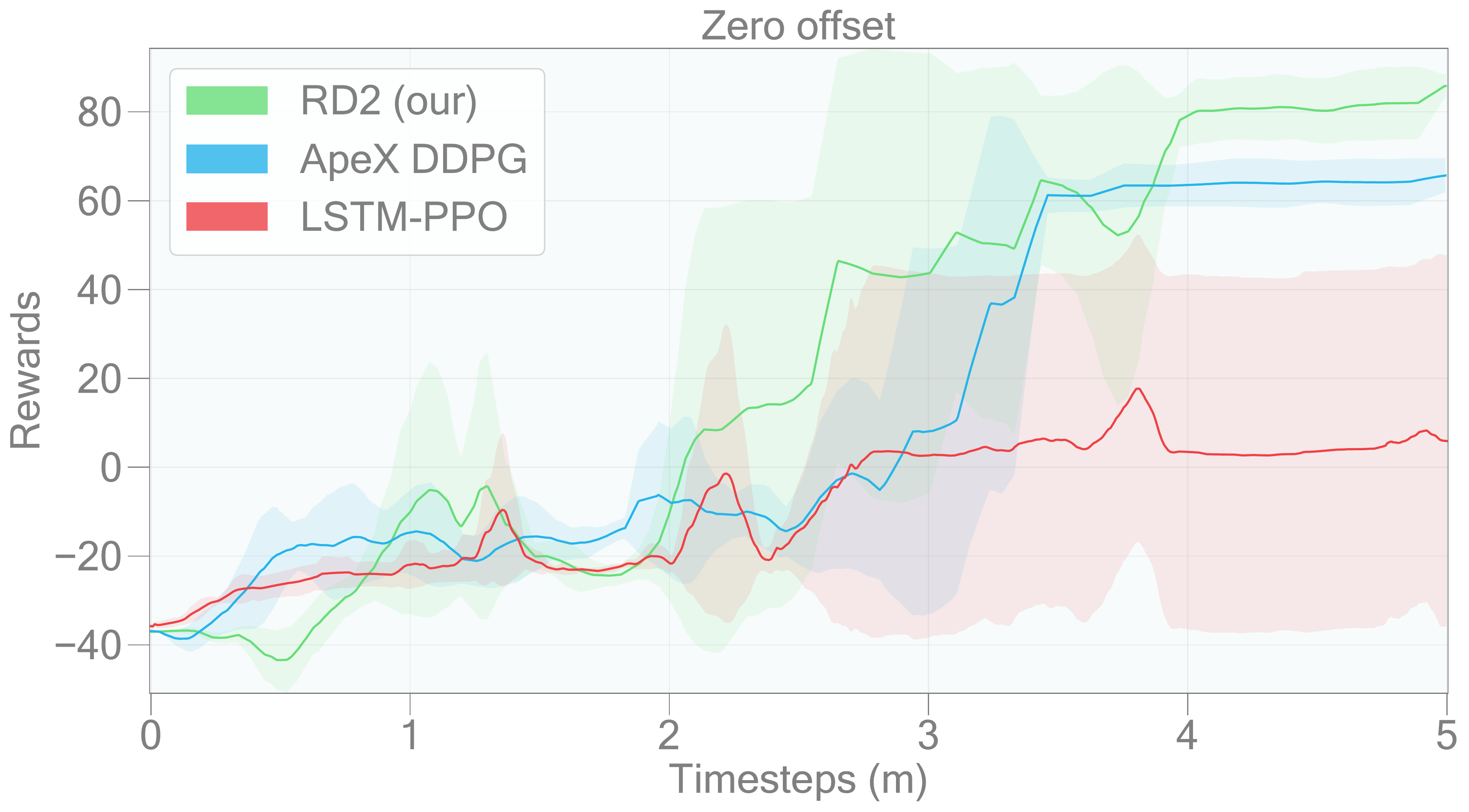}} & \includegraphics[width=0.31\textwidth]{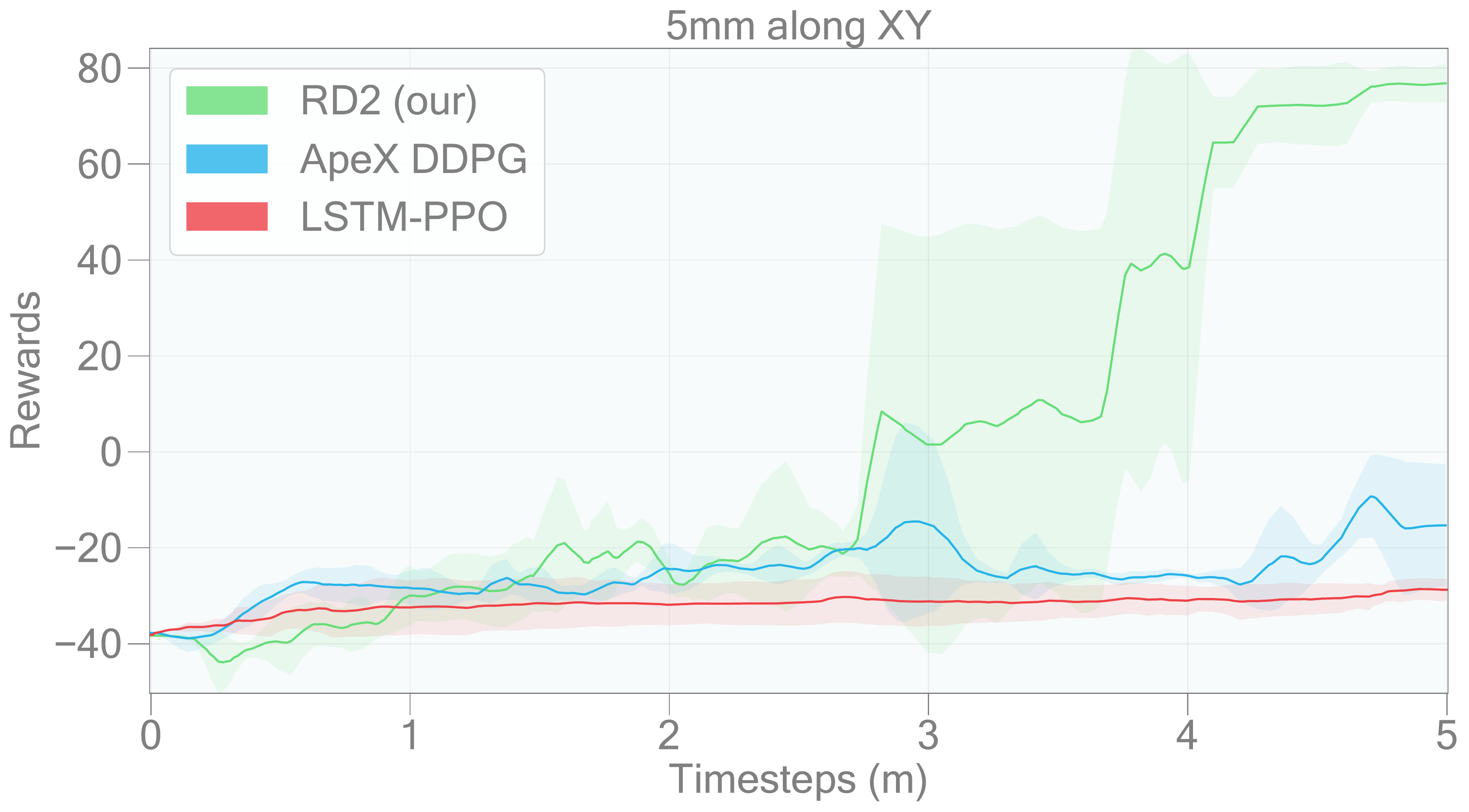} & \includegraphics[width=0.31\textwidth]{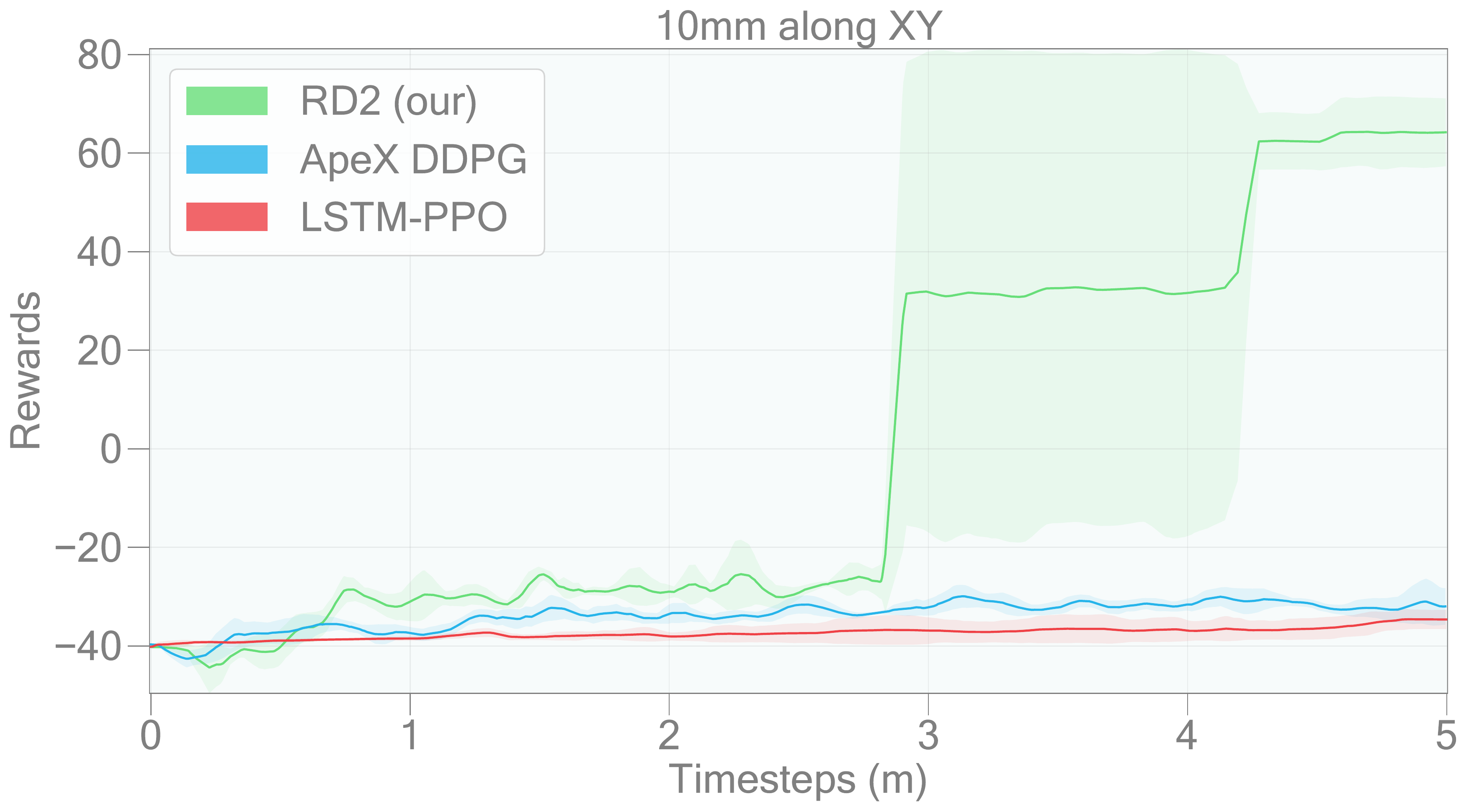} \\
         &\includegraphics[width=0.31\textwidth]{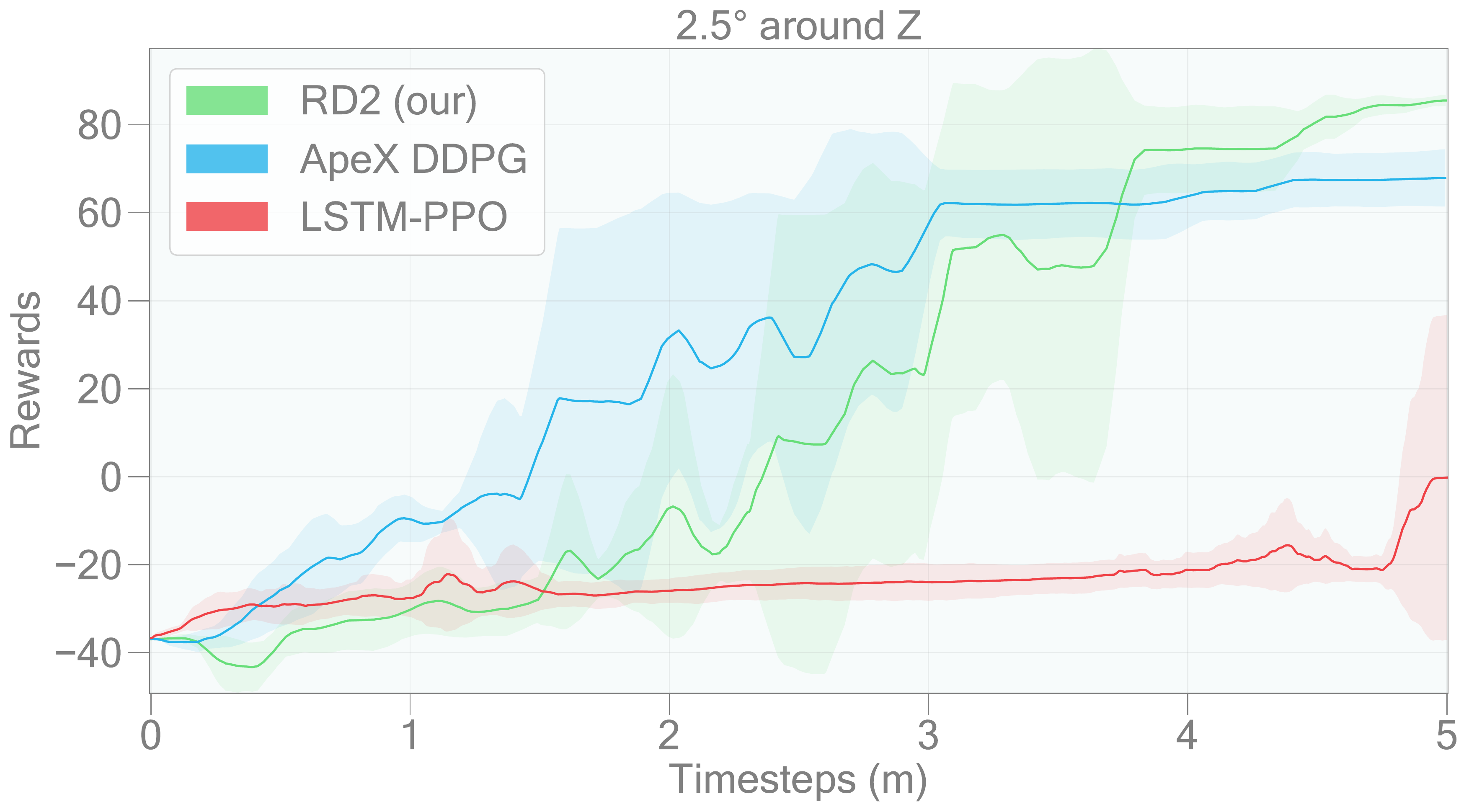} & \includegraphics[width=0.31\textwidth]{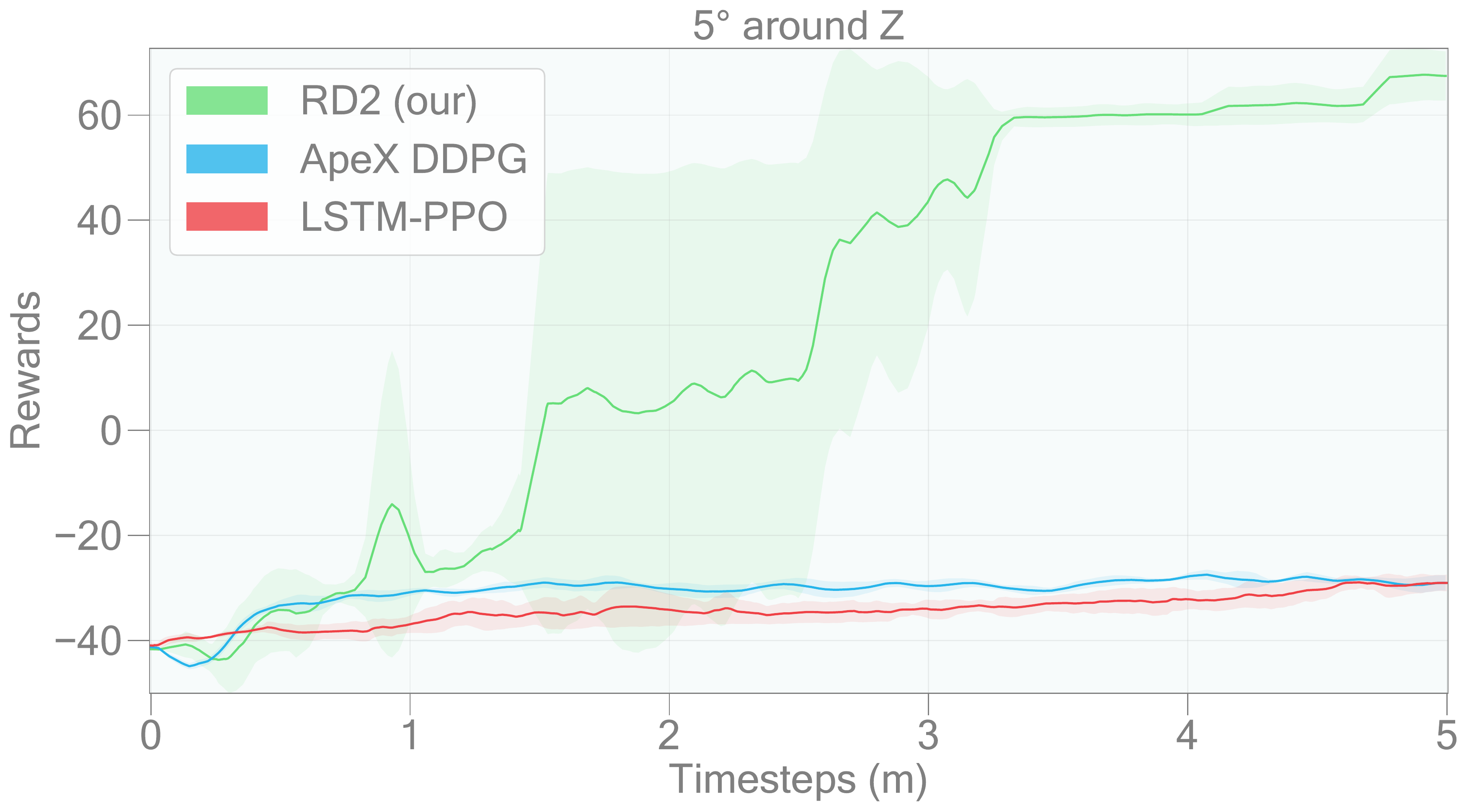}\\
    \end{tabular}
    \caption{Rewards comparison for the \textbf{lap-joint} tasks. The difficulty of the tasks increases from left to right. The lines visualize the average of the best model performance across time for three PBT runs with different random seeds and the shaded areas show the 95\% confidence bound.}
    \label{fig:lap-joint-results}
\end{figure*}

\begin{figure*}[h]
    \centering
    \setlength{\tabcolsep}{0pt}
    \begin{tabular}{ c c c }
         \multirow{2}{*}[3em]{\includegraphics[width=0.31\textwidth]{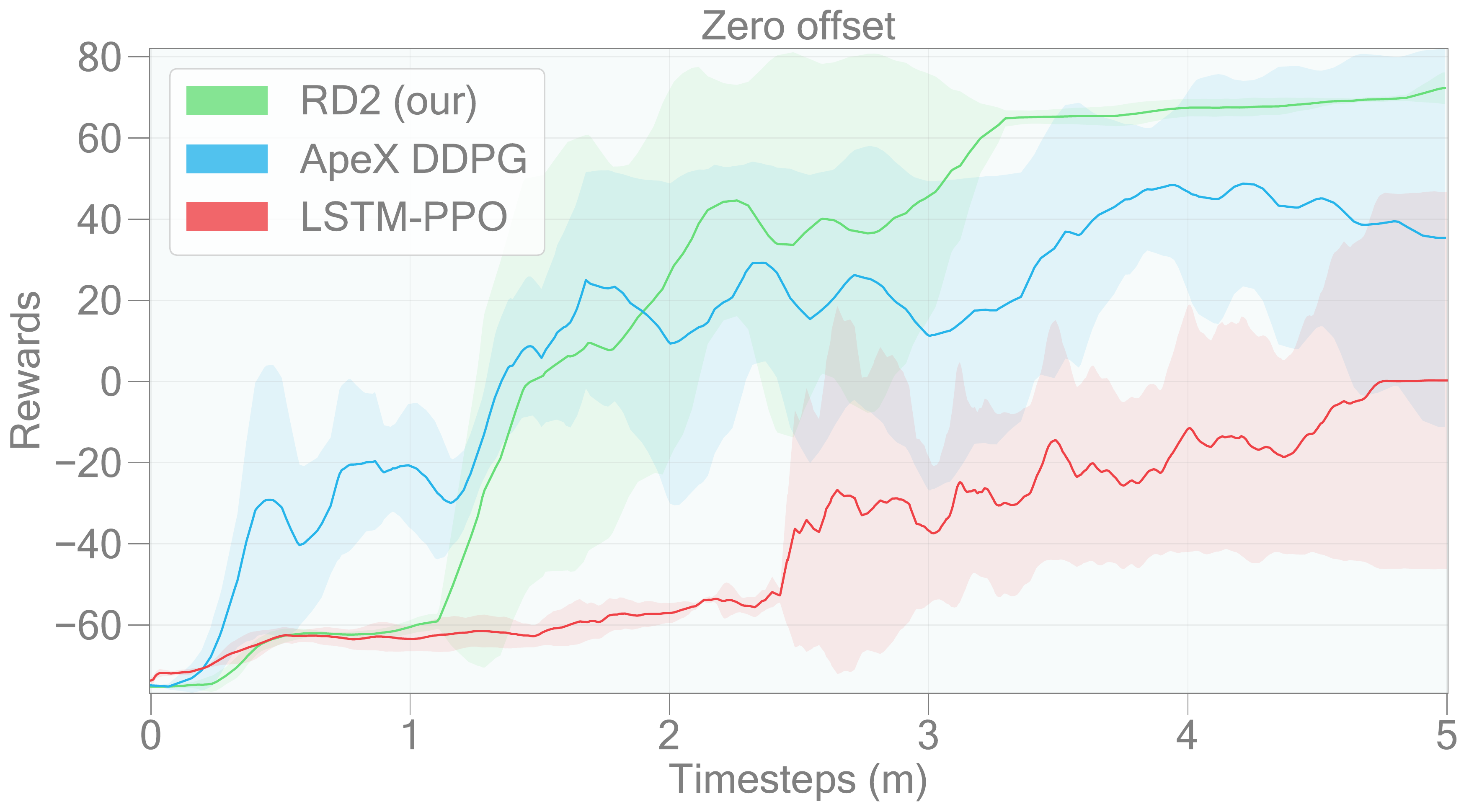}} & \includegraphics[width=0.31\textwidth]{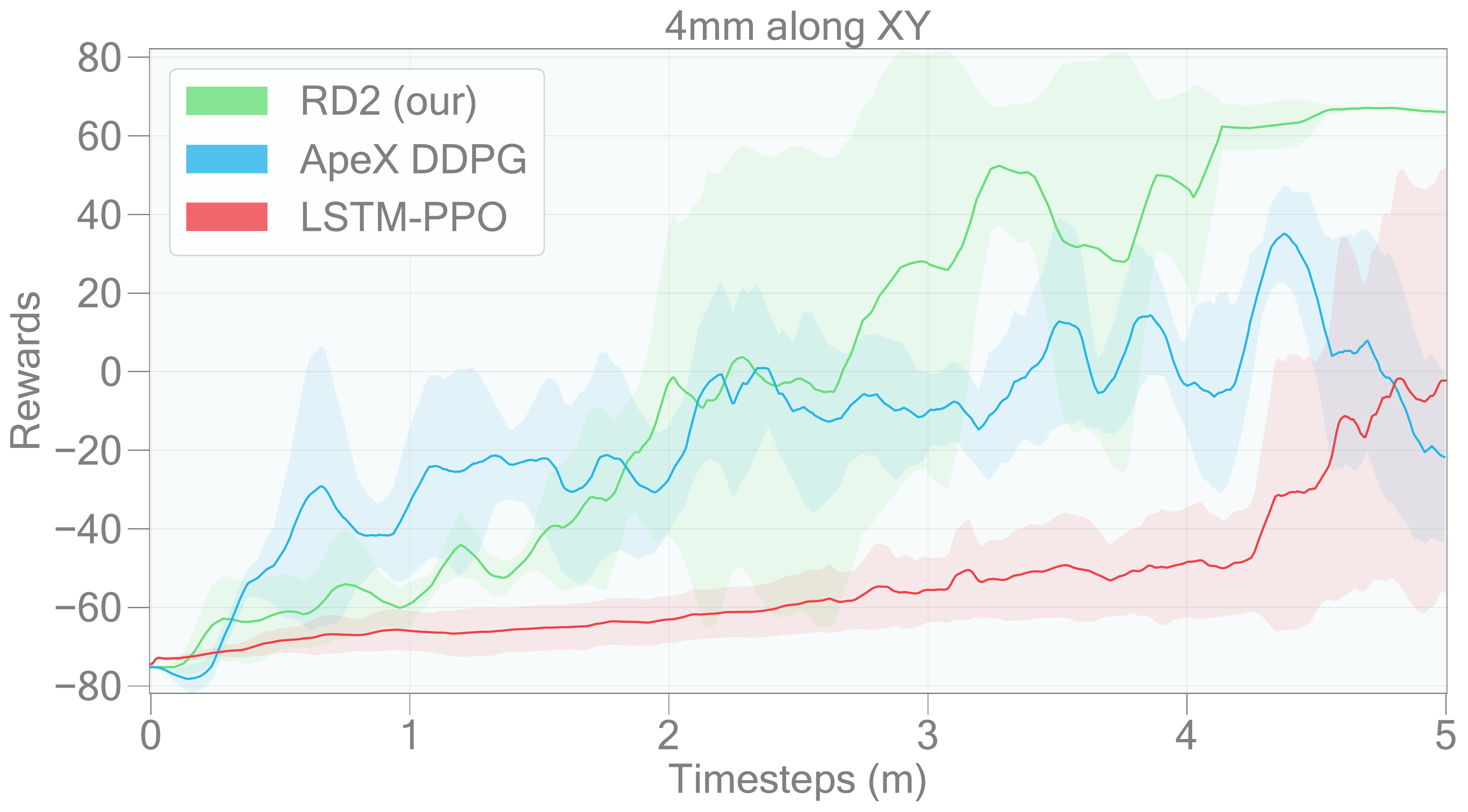} & \includegraphics[width=0.31\textwidth]{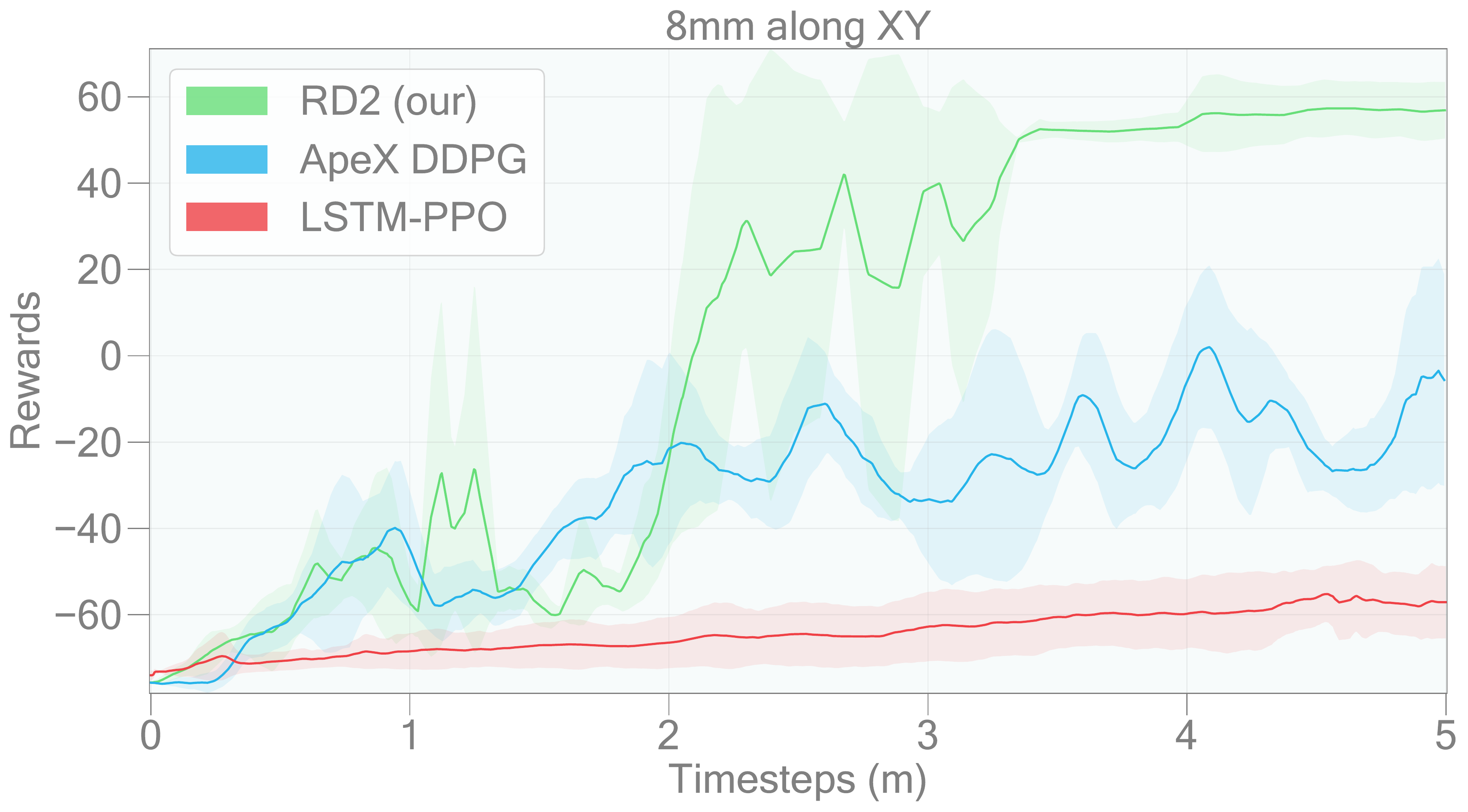} \\
         &\includegraphics[width=0.31\textwidth]{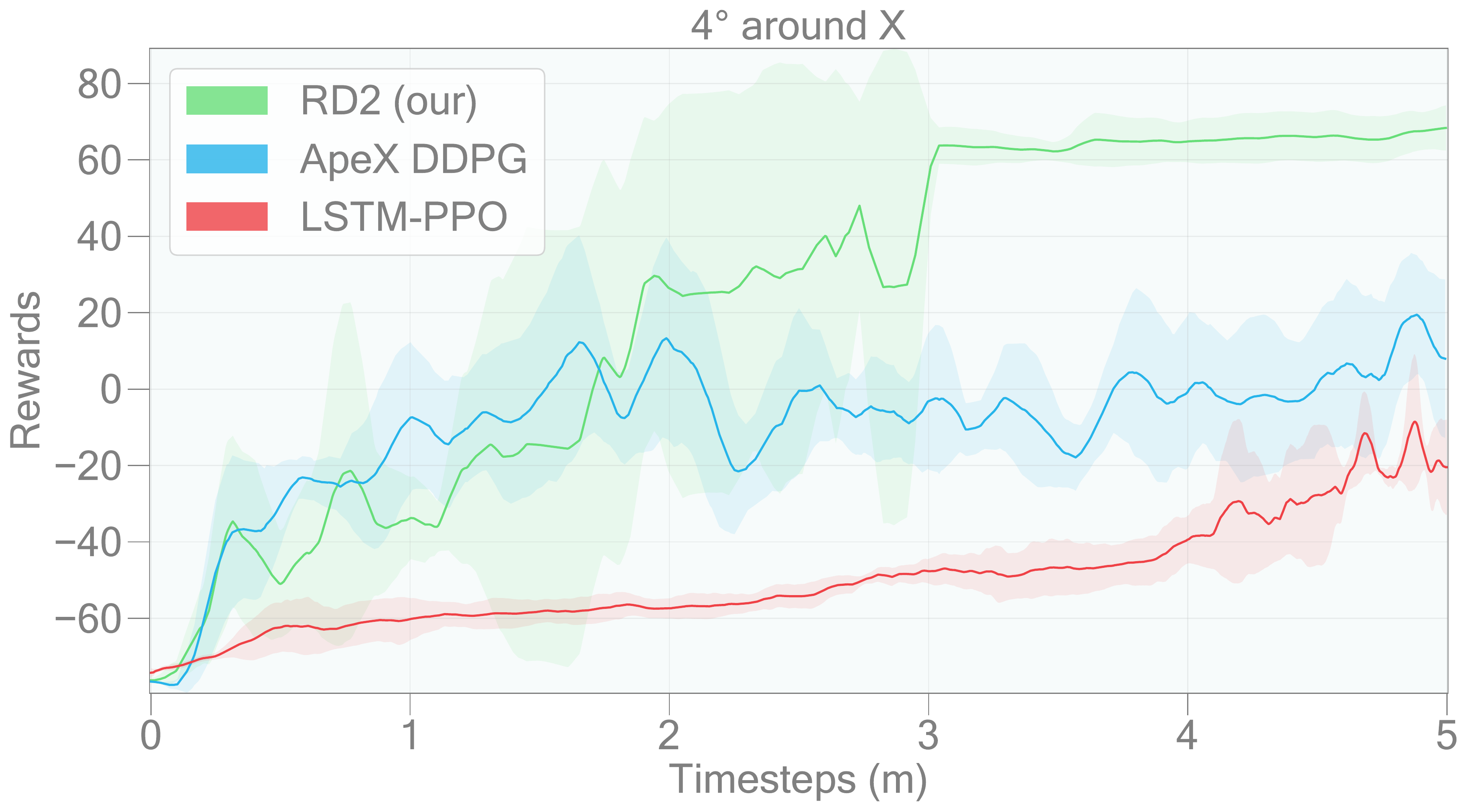} & \includegraphics[width=0.31\textwidth]{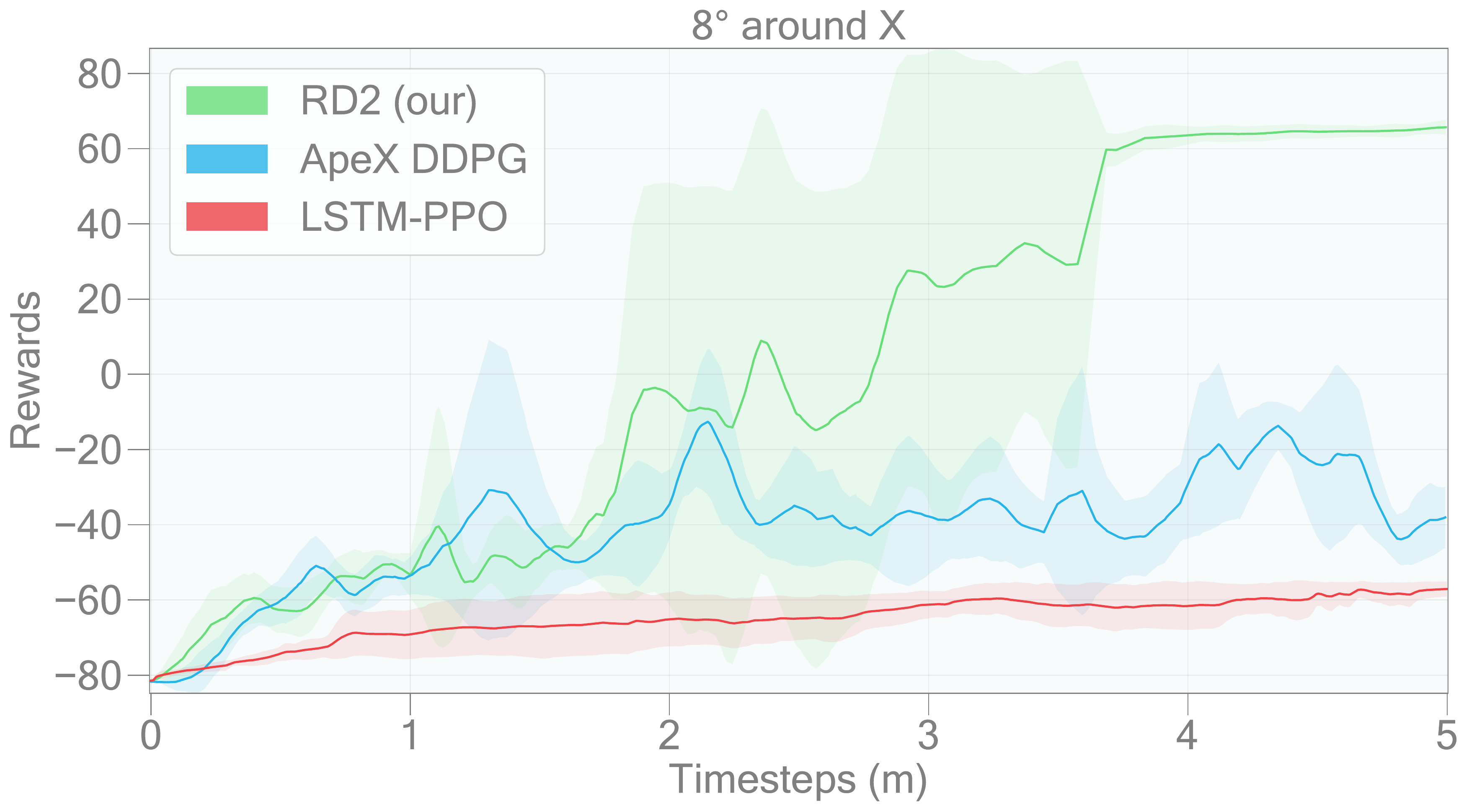}\\
    \end{tabular}
    \caption{Rewards comparison for the \textbf{peg-in-hole} tasks. The difficulty of the tasks increases from left to right. The lines visualize the average of the best model performance across time for three PBT runs with different random seeds and the shaded areas show the 95\% confidence bound.}
    \label{fig:peg-in-hole-results}
\end{figure*}

To answer these questions, we create two customized robotless assembly environments, \textbf{lap-joint} and \textbf{peg-in-hole}, to evaluate the performance of RD2 in comparison to SOTA RL algorithms in the continuous action domain. To access the adaptability of the trained robot-agnositc polices, we rollout the policies on three robotic arms, Franka Panda, Kuka KR60, and UR10, with varying initial offsets and physical noise injected in simulation. We use an internal simulator that simulates the available robots in our lab and contains a variety of embedded drivers of the physical hardware \footnote{The internal simulator is built for easy deployment of sim-trained policies on physical robots, which unfortunately due to COVID-19 we have not been able to access.}. The video presenting our experiments is available at \url{https://sites.google.com/view/rd2-rl}

\subsection{Robotless Assembly Environments}
\label{robotless-envs}

In the training environment in simulation, both the lap-joint and peg-in-hole tasks are performed without robots. For each environment, we only include objects contributing to F/T readings. Specifically, the lap-joint environment consists of a model of a customized gripper, of a sensor, and of a pair of joint members, shown in Fig.~\ref{fig:assembly-env}(left); the peg-in-hole environment consists of a model of the Franka Panda gripper, of a sensor, of a tampered peg and a hole, shown in Fig.~\ref{fig:assembly-env}(right). Every single dynamic object is assigned an estimated inertial property (mass and centre of mass) and applied friction. The F/T sensors are gravity compensated. The tolerance of the lap-joint task is 2mm and the tolerance of the peg-in-hole task is 0mm.

\subsection{RD2 Performance}  
\label{results}

We design 5 tasks for each environment with varying complexity, by setting the gripper at different initial poses, so the joint member or the peg on the gripper has an angular offset or a linear offset from its default pose. In general, the larger the offset the more difficult it is to train. We compare RD2 to (1) Ape-X DDPG, which has the similar architecture of RD2 but without memory augmentation and the replay buffer improvements in RD2, (2) LSTM-PPO, which performs on-policy training with memory augmentation. \footnote{Same as RD2, the two algorithms also trained with PBT with 8 concurrent trials, each of which contains one single GPU-based learner and 8 actors.} We plot the average reward reached by the agent against the number of timesteps of training for each algorithm, as shown in Fig.\ref{fig:lap-joint-results} for the lap-joint tasks and Fig.\ref{fig:peg-in-hole-results} for the peg-in-hole tasks. Note that positive reward in the plots indicates successful assemblies and the higher the positive reward, the higher the success rate there is for the assembly tasks.

Both figures show that RD2 outperforms the other two algorithms across all the tasks. As the difficulty increases in each environment, RD2 keeps a stable performance while the performance of the other two algorithms drop significantly. In general, the off-policy algorithms (RD2 and Ape-X DDPG) perform better than the on-policy algorithm (LSTM-PPO) on our tasks, which suggests the importance of sample efficiency for partially observable environments. However, the declined performance of Ape-X DDPG on harder tasks in both environments indicates the necessity of our improvements on RD2, especially in real-world settings where small misalignments are inevitable. 




\subsection{Evaluation of Adaptation}
\label{eval}
In this section, we evaluate how well the robotless policies transfer to different robotic arms and how they generalize to different initial pose offsets and different physical noises. Specifically, we take into account the following factors: linear and angular offset, and Gaussian noise in F/T measurements and in friction.  

We use the lap-joint task environment for evaluating different initial pose offsets. Table~\ref{table: robot-eval} shows the result of evaluating the same robotless policy on three different robotic arms, with 0 offset and with position and orientation offsets. We show the success rate over 10 runs. The trained robotless policy transfers well to different robots and it generalizes well to different initial position and orientation offsets for the Panda and UR10 robots. In future work we will investigate in the implementation of our internal simulator why the policy does not generalize as well on the Kuka robot.

\begin{table}[h]
\setlength{\arrayrulewidth}{0.1mm}
\setlength{\tabcolsep}{2pt}
\renewcommand{\arraystretch}{1.6}
\centering
\begin{tabular}{|c|c|c|c|}
 \hline
 \textbf{Robots} & \textbf{0 offset} & \textbf{3mm offset along XYZ} & \textbf{5 deg offset around Z}\\
 \hline
 Franka Panda & 100\% & 100\% & 100\% \\
 \hline
 UR10 & 100\% & 90\% & 90\% \\
 \hline
 Kuka KR60 & 100\% & 40\% & 30\% \\
 \hline
\end{tabular}
\caption{We show the success rate of 10 rollouts of the same robotless policy of the lap-joint task on three different robotic arms: Panda, UR10 and Kuka. The same policy is also tested with varying initial position and orientation offsets}
\label{table: robot-eval}
\end{table}

We use the peg-in-hole task environment for evaluating physical noises. Table~\ref{table: noise-eval} shows the result of evaluating the same robotless policy on three different robotic arms, with no noise and with Gaussian noise in F/T measurements and friction. We show the success rate over 10 runs. The policy generalizes well when Gaussian noise of $0$ mean and $20\%$ variance added to F/T measurements and to friction.
\begin{table}[h]
\setlength{\arrayrulewidth}{0.1mm}
\setlength{\tabcolsep}{2pt}
\renewcommand{\arraystretch}{1.6}
\centering
\begin{tabular}{|c|c|c|c|}
 \hline
 \textbf{Robots} & \textbf{0 noise} & \textbf{20\% F/T noise} & \textbf{20\% friction noise}\\
 \hline
 Franka Panda & 100\% & 80\% & 100\% \\
 \hline
 UR10 & 100\% & 80\% & 90\% \\
 \hline
 Kuka KR60 & 100\% & 90\% & 100\% \\
 \hline
\end{tabular}
\caption{We show the success rate of 10 rollouts of the same robotless policy of the peg-in-hole task on three different robotic arms: Panda, UR10 and Kuka. The policy is tested with Gaussian noise of $0$ mean and $20\%$ variance added to F/T, as well as with Gaussian noise of $0$ mean and $20\%$ variance added to friction.}
\label{table: noise-eval}
\end{table}

\begin{figure}[h]
    \centering
    \includegraphics[width=0.48\textwidth]{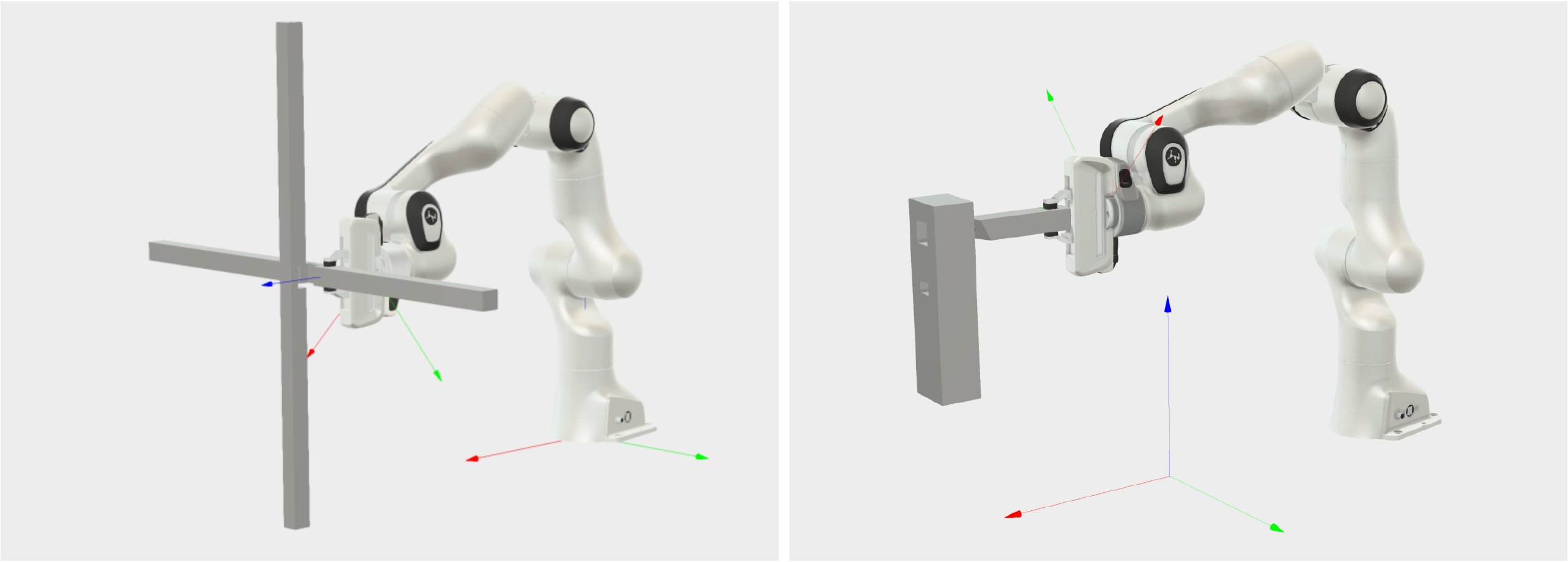}
    \caption{Visualizations of evaluating the trained policies on assembly pieces at different poses from the training pose.}
    \label{fig:rotation-eval}
\end{figure}

We further evaluate the policies on assembly pieces that are placed at different poses as opposed to the training pose, as shown in Fig.\ref{fig:rotation-eval}. As it is common to have the same assembly at various poses during construction, it is important for a trained policy to be able to generalize to different poses. For each task, when we transform the actions to be in the frame of the target during training, our trained policy can successfully transfer to tasks with varying target poses.

\section{Conclusion and Future Work}
\label{sec:conclusion}

This paper presented a learning approach to solving high-precision robotic assembly tasks with the focus on the contact-rich phase. We use F/T measurements as the only observation. The approach allows different robots to share one trained policy and to adapt to misalignment without requiring external position tracking or vision systems, which improves the capability and scalability of RL-based robotic systems in unstructured settings, such as architectural construction. To achieve it, RD2 learns a memory-based representation to compensate for partial observability. Training takes place in robotless environments and trained polices can transfer to various robotic arms without re-training. Our results show that RD2 achieves the best performance on all assembly tasks in comparison to two baselines, Ape-X DDPG and LSTM-PPO. Furthermore, RD2 demonstrates its strength on harder tasks that cannot be solved by either baseline algorithm. We also show that our trained policies transfer well to different robotic arms, and can adapt to various initial pose misalignment as well as noise injected in F/T measurements and friction parameters.

Future work includes comparing our approach to augmenting F/T measurements with vision as observations. When hardware access becomes possible, we will deploy the trained polices on different physical robots to further assess the approach.  
 
Although the reward function is solely used during training in simulation, where distance is easy to acquire, and not used during rollouts, we suspect using distance in the reward function may be detrimental to the adaptability of trained policies. We will investigate reward learning in the future.

\section*{APPENDIX}

\subsection{The hyper-parameters fine-tuned in RD2 using PBT}
\begin{table}[H]
    \setlength{\arrayrulewidth}{0.1mm}
    \setlength{\tabcolsep}{8pt}
    \renewcommand{\arraystretch}{1.6}
    \centering
    \footnotesize	
    \begin{tabular}{|c|c|}
     \hline
     \textbf{Fine-tuned Hyperparameter} & \textbf{Fine-tuned range}\\
     \hline
     Number of batches & (20, 120)$^1$ \\
     \hline
     Sequence length & [16, 32, 64, 128]$^2$  \\
     \hline
     Target network update frequency & [25000, 50000, 75000, 100000] \\
     \hline
    N-step & (3, 8) \\
    \hline
    Minimal iteration time (s) & [30, 40, 50, 60] \\
     \hline
    \end{tabular}
    \\$^1$ Sampled from the given range, $^2$ Sampled from the given list
    \caption{The hyper-parameters fine-tuned in RD2 using PBT}
    \label{table: tasks}
\end{table}


\section*{ACKNOWLEDGMENT}

We thank Tonya Custis and Erin Bradner for budgetary support of the project; Yotto Koga for the development of the simulator to run our experiments; Pantelis Katsiaris for the helpful discussions.


\bibliographystyle{./bibliography/IEEEtran}
\bibliography{reference.bib}

\begin{thebibliography}{10}
\providecommand{\url}[1]{#1}
\csname url@samestyle\endcsname
\providecommand{\newblock}{\relax}
\providecommand{\bibinfo}[2]{#2}
\providecommand{\BIBentrySTDinterwordspacing}{\spaceskip=0pt\relax}
\providecommand{\BIBentryALTinterwordstretchfactor}{4}
\providecommand{\BIBentryALTinterwordspacing}{\spaceskip=\fontdimen2\font plus
\BIBentryALTinterwordstretchfactor\fontdimen3\font minus
  \fontdimen4\font\relax}
\providecommand{\BIBforeignlanguage}[2]{{%
\expandafter\ifx\csname l@#1\endcsname\relax
\typeout{** WARNING: IEEEtran.bst: No hyphenation pattern has been}%
\typeout{** loaded for the language `#1'. Using the pattern for}%
\typeout{** the default language instead.}%
\else
\language=\csname l@#1\endcsname
\fi
#2}}
\providecommand{\BIBdecl}{\relax}
\BIBdecl

\bibitem{horgan2018distributed}
D.~Horgan, J.~Quan, D.~Budden, G.~Barth-Maron, M.~Hessel, H.~Van~Hasselt, and
  D.~Silver, ``Distributed prioritized experience replay,'' \emph{arXiv
  preprint arXiv:1803.00933}, 2018.

\bibitem{schaul2015prioritized}
T.~Schaul, J.~Quan, I.~Antonoglou, and D.~Silver, ``Prioritized experience
  replay,'' \emph{arXiv preprint arXiv:1511.05952}, 2015.

\bibitem{zhu2020ingredients}
H.~Zhu, J.~Yu, A.~Gupta, D.~Shah, K.~Hartikainen, A.~Singh, V.~Kumar, and
  S.~Levine, ``The ingredients of real-world robotic reinforcement learning,''
  \emph{arXiv preprint arXiv:2004.12570}, 2020.

\bibitem{andrychowicz2020learning}
O.~M. Andrychowicz, B.~Baker, M.~Chociej, R.~Jozefowicz, B.~McGrew,
  J.~Pachocki, A.~Petron, M.~Plappert, G.~Powell, A.~Ray \emph{et~al.},
  ``Learning dexterous in-hand manipulation,'' \emph{The International Journal
  of Robotics Research}, vol.~39, no.~1, pp. 3--20, 2020.

\bibitem{ren2018learning}
T.~Ren, Y.~Dong, D.~Wu, and K.~Chen, ``Learning-based variable compliance
  control for robotic assembly,'' \emph{Journal of Mechanisms and Robotics},
  vol.~10, no.~6, p. 061008, 2018.

\bibitem{schoettler2019deep}
G.~Schoettler, A.~Nair, J.~Luo, S.~Bahl, J.~A. Ojea, E.~Solowjow, and
  S.~Levine, ``Deep reinforcement learning for industrial insertion tasks with
  visual inputs and natural rewards,'' \emph{arXiv preprint arXiv:1906.05841},
  2019.

\bibitem{akkaya2019solving}
I.~Akkaya, M.~Andrychowicz, M.~Chociej, M.~Litwin, B.~McGrew, A.~Petron,
  A.~Paino, M.~Plappert, G.~Powell, R.~Ribas \emph{et~al.}, ``Solving rubik's
  cube with a robot hand,'' \emph{arXiv preprint arXiv:1910.07113}, 2019.

\bibitem{luo2020dynamic}
J.~Luo and H.~Li, ``Dynamic experience replay,'' \emph{arXiv preprint
  arXiv:2003.02372}, 2020.

\bibitem{beltran2020variable}
C.~C. Beltran-Hernandez, D.~Petit, I.~G. Ramirez-Alpizar, and K.~Harada,
  ``Variable compliance control for robotic peg-in-hole assembly: A
  deep-reinforcement-learning approach,'' \emph{Applied Sciences}, vol.~10,
  no.~19, p. 6923, 2020.

\bibitem{barth2018distributed}
G.~Barth-Maron, M.~W. Hoffman, D.~Budden, W.~Dabney, D.~Horgan, A.~Muldal,
  N.~Heess, and T.~Lillicrap, ``Distributed distributional deterministic policy
  gradients,'' in \emph{6th International Conference on Learning
  Representations}, 2018.

\bibitem{lillicrap2016ddpg}
T.~P. Lillicrap, J.~J. Hunt, A.~Pritzel, N.~Heess, T.~Erez, Y.~Tassa,
  D.~Silver, and D.~Wierstra, ``Continuous control with deep reinforcement
  learning,'' in \emph{6th International Conference on Learning
  Representations}, 2016.

\bibitem{heess2015memory}
N.~Heess, J.~J. Hunt, T.~P. Lillicrap, and D.~Silver, ``Memory-based control
  with recurrent neural networks,'' \emph{arXiv preprint arXiv:1512.04455},
  2015.

\bibitem{schulman2017proximal}
J.~Schulman, F.~Wolski, P.~Dhariwal, A.~Radford, and O.~Klimov, ``Proximal
  policy optimization algorithms,'' \emph{arXiv preprint arXiv:1707.06347},
  2017.

\bibitem{mnih2015dqn}
V.~Mnih, K.~Kavukcuoglu, D.~Silver, A.~A. Rusu, J.~Veness, M.~G. Bellemare,
  A.~Graves, M.~Riedmiller, A.~K. Fidjeland, G.~Ostrovski \emph{et~al.},
  ``Human-level control through deep reinforcement learning,'' \emph{Nature
  518, pages 529–533}, 2015.

\bibitem{espeholt2018impala}
L.~Espeholt, H.~Soyer, R.~Munos, K.~Simonyan, V.~Mnih, T.~Ward, Y.~Doron,
  V.~Firoiu, T.~Harley, I.~Dunning \emph{et~al.}, ``Impala: Scalable
  distributed deep-rl with importance weighted actor-learner architectures,''
  \emph{arXiv preprint arXiv:1802.01561}, 2018.

\bibitem{fan2018guided}
Y.~Fan, J.~Luo, and M.~Tomizuka, ``A learning framework for high precision
  industrial assembly,'' \emph{arXiv preprint arXiv:1809.08548v3}, 2018.

\bibitem{levine2013guided}
S.~Levine and V.~Koltun, ``Guided policy search,'' in \emph{International
  Conference on Machine Learning}, 2013, pp. 1--9.

\bibitem{sutton2018reinforcement}
R.~S. Sutton and A.~G. Barto, \emph{Reinforcement learning: An
  introduction}.\hskip 1em plus 0.5em minus 0.4em\relax MIT press, 2018.

\bibitem{lee2019making}
M.~A. Lee, Y.~Zhu, K.~Srinivasan, P.~Shah, S.~Savarese, L.~Fei-Fei, A.~Garg,
  and J.~Bohg, ``Making sense of vision and touch: Self-supervised learning of
  multimodal representations for contact-rich tasks,'' in \emph{2019
  International Conference on Robotics and Automation (ICRA)}.\hskip 1em plus
  0.5em minus 0.4em\relax IEEE, 2019, pp. 8943--8950.

\bibitem{li2021}
A.~Apolinarska*, M.~Pacher*, H.~Li*, N.~Cote, R.~Pastrana, F.~Gramazio, and
  M.~Kohler, ``Robotic assembly of timber joints using reinforcement
  learning,'' \emph{Automation in Construction Journal,
  https://doi.org/10.1016/j.autcon.2021.103569}, vol. 125, 2021.

\bibitem{hausknecht2015deep}
M.~Hausknecht and P.~Stone, ``Deep recurrent q-learning for partially
  observable mdps,'' in \emph{2015 AAAI Fall Symposium Series}, 2015.

\bibitem{kapturowski2018recurrent}
S.~Kapturowski, G.~Ostrovski, J.~Quan, R.~Munos, and W.~Dabney, ``Recurrent
  experience replay in distributed reinforcement learning,'' in
  \emph{International conference on learning representations}, 2018.

\bibitem{openai2018learning}
\BIBentryALTinterwordspacing
OpenAI, M.~Andrychowicz, B.~Baker, M.~Chociej, R.~Józefowicz, B.~McGrew,
  J.~Pachocki, A.~Petron, M.~Plappert, G.~Powell, A.~Ray, J.~Schneider,
  S.~Sidor, J.~Tobin, P.~Welinder, L.~Weng, and W.~Zaremba, ``Learning
  dexterous in-hand manipulation,'' \emph{CoRR}, 2018. [Online]. Available:
  \url{http://arxiv.org/abs/1808.00177}
\BIBentrySTDinterwordspacing

\bibitem{inoue2017deep}
T.~Inoue, G.~De~Magistris, A.~Munawar, T.~Yokoya, and R.~Tachibana, ``Deep
  reinforcement learning for high precision assembly tasks,'' in \emph{2017
  IEEE/RSJ International Conference on Intelligent Robots and Systems
  (IROS)}.\hskip 1em plus 0.5em minus 0.4em\relax IEEE, 2017, pp. 819--825.

\bibitem{coumans2016pybullet}
E.~Coumans and Y.~Bai, ``Pybullet, a python module for physics simulation for
  games, robotics and machine learning,'' \emph{GitHub repository}, 2016.

\bibitem{werbos1990backpropagation}
P.~J. Werbos, ``Backpropagation through time: what it does and how to do it,''
  \emph{Proceedings of the IEEE}, vol.~78, no.~10, pp. 1550--1560, 1990.

\bibitem{jaderberg2017population}
M.~Jaderberg, V.~Dalibard, S.~Osindero, W.~M. Czarnecki, J.~Donahue, A.~Razavi,
  O.~Vinyals, T.~Green, I.~Dunning, K.~Simonyan \emph{et~al.}, ``Population
  based training of neural networks,'' \emph{arXiv preprint arXiv:1711.09846},
  2017.

\end{thebibliography}

\end{document}